\documentclass[runningheads]{llncs}
\usepackage[T1]{fontenc}
\usepackage{booktabs}
\usepackage{graphicx}
\usepackage{subcaption}
\begin{document}
\title{Distributed Online Life-Long Learning (DOL3) for Multi-agent Trust and Reputation Assessment in E-commerce}
\titlerunning{DOL3 for MAS TRA}
% If the paper title is too long for the running head, you can set
% an abbreviated paper title here
%
\author{Hariprasauth Ramamoorthy\orcidID{0009-0004-4922-0319} \and
Shubhankar Gupta\orcidID{0009-0009-2028-9992} \and
Suresh Sundaram\orcidID{0000-0001-6275-0921}}
\authorrunning{Ramamoorthy et al.}
% First names are abbreviated in the running head.
% If there are more than two authors, 'et al.' is used.
%
\institute{Indian Institute of Science, Bengaluru, Karnataka, India
\email{\{hariprasauth,shubhankarg,vssuresh\}@iisc.ac.in}\\
\url{https://iisc.ac.in}}
\maketitle              % typeset the header of the contribution
\begin{abstract}
Trust and Reputation Assessment of service providers in citizen-focused environments like e-commerce is vital to maintain the integrity of the interactions among agents. The goals and objectives of both the service provider and service consumer agents are relevant to the goals of the respective citizens (end users). The provider agents often pursue selfish goals that can make the service quality highly volatile, contributing towards the non-stationary nature of the environment. The number of active service providers tends to change over time resulting in an open environment. This necessitates a rapid and continual assessment of the Trust and Reputation. A large number of service providers in the environment require a distributed multi-agent Trust and Reputation assessment. This paper addresses the problem of multi-agent Trust and Reputation Assessment in a non-stationary environment involving transactions between providers and consumers. In this setting, the observer agents carry out the assessment and communicate their assessed trust scores with each other over a network. We propose a novel Distributed Online Life-Long Learning (DOL3) algorithm that involves real-time rapid learning of trust and reputation scores of providers. Each observer carries out an adaptive learning and weighted fusion process combining their own assessment along with that of their neighbour in the communication network. Simulation studies reveal that the state-of-the-art methods, which usually involve training a model to assess an agent's trust and reputation, do not work well in such an environment. The simulation results show that the proposed DOL3 algorithm outperforms these methods and effectively handles the volatility in such environments. From the statistical evaluation, it is evident that DOL3 performs better compared to other models in 90\% of the cases. 

\end{abstract}
\section{Introduction}
Multi-agent systems (MAS) in Distributed Artificial Intelligence (DAI) have the capability to address complex computing problems in Computer Science, Civil Engineering, Robotics, Economics, etc.; see, for instance \cite{Dorri}. The agents in such an architecture use their knowledge autonomously to decide and act in their environment \cite{Shamshirband}. One of the major use cases for MAS is in the area of e-commerce, where the agents are distributed in an environment and act autonomously towards their goals, playing various roles like a negotiator, buyer, service provider, consumer, etc \cite{M. Tomášek and J. Trelová}. In e-commerce, the agents widely play the role of either a Service Provider or a Service Consumer. The agents act as the representatives of the users at the service provider and service consumer. 

In most e-commerce scenarios, the provider would act selfishly to gain the consumer's trust to improve their reputation among the consumers. \cite{Jășcanu} introduced a novel approach to model this behavior for the service providers by attaching emotional quotients to their interactions. Trust and reputation in such scenarios play a vital role in assisting consumers in identifying the providers to choose from. Adding to the complexity is the noisy data that impacts the way the multi-agents understand the system \cite{Zhang}. Several case studies including those in \cite{Wu} talk about how malicious sellers deceive and manipulate the viewers. The decentralized marketplace provides better filter and search mechanisms thereby introducing more autonomy for the agents in the interactions \cite{Jan Tscheke et al.}, further highlighting the importance of Trust and Reputation assessment in such scenarios.  

\subsection{Contribution}
In this paper, we extend the MAS architecture defined in \cite{Ehikioya} with an observer agent to perform the Trust and Reputation Assessment of service providers. The provider's quality of service can be highly volatile. The incorrect learning during multi-agent interactions leads to a risk that would show infectious growth as agents interact and learn from each other \cite{Chelarescu}.

In this paper, we propose a novel Distributed Online Life-Long Learning (DOL3) framework that involves the online learning of trust and reputation scores of service providers by a set of observers communicating their opinions with each other. Each observer runs the DOL3 algorithm in a decentralized manner. The DOL3 algorithm involves an adaptive online learning framework coupled with a trust fusion process, effectively combining an observer's assessment with its neighboring observers in the interaction network. The online learning process in the DOL3 algorithm is inspired by that of the exponentially weighted online learning forecaster \cite{Cesa}. Simulation results show that DOL3 outperforms the state-of-the-art machine learning assessment methods; such machine learning methods usually involve training a machine learning model to assess an agent’s trust and reputation in a stationary environment. On the other hand, owing to its rapid online learning capability, DOL3 deals with the non-stationary environment effectively. 

For the simulation studies, different types of social networks have been considered that are essential to understand how the agents are wired to interact with each other and illustrate the social (network) connections among the agents, as stated in \cite{Dimitri}. The three networks - Small world, Scale-free, and Regular networks with Homophily are considered during the simulation to understand how the DOL3 algorithm performs compared to the other methods. To perform the statistical evaluation of these findings, we applied the comparison with real-world data - Movie recommendation system \cite{Goyani}, for which the data set was taken from \cite{kaggle}.

Simulations involve some recommendation agents becoming malicious in an intermittent fashion. The recommendation system agents are evaluated by the observer agents to help the users get the right list of movies. This environment setup is used to evaluate how the DOL3 algorithm in various network types with different parameters performs compared to that of other state-of-the-art methods and it is evident that the DOL3 algorithm performs better compared to other state-of-the-art models in 90\% of the cases.

\section{Distributed Online Life-Long Learning (DOL3)}
\subsection{Problem formulation} \label{problemformulation}
The multi-agent architecture considered in this paper involves three types of citizen-centric agents: service providers, consumers, and observers. The edges (connecting lines) between observers and providers indicate that those specific observers have visibility limited to the linked providers. The edges among the observers indicate their neighborhood where the information sharing happens. The consumers can interact with only those providers that they are connected to as per the interaction network. An observer is tasked to do a timely and effective assessment of the providers' quality of service to guide the consumers with the best possible provider. 

Let there be $N_p$ service providers, $N_o$ observers, and $N_c$ consumers. Let $\Omega_i$ denote the set containing indices of all the providers that are observed by the $i^{th}$ observer as per an interaction network $G$. Denote $\Lambda_i$ as the set containing the indices of all the observers that are neighbors to the $i^{th}$ observer as per the interaction network $G$. Further, let $\Gamma_i$ be the set of consumers that receive recommendations from the $i^{th}$ observer as per the interaction network $G$. 

It is assumed that the consumers can purchase services one by one w.r.t. interaction count $t$, with only one consumer per interaction count, i.e., $1^{st}$ consumer purchases at $t=1$, $2^{nd}$ consumer purchases at $t=2$, and so on, such that the $i^{th}$ consumer purchases only at the interaction counts given by the count sequence: $t_{n,i} = (n-1) \cdot N_c + i$, where $n = 1,2,\cdots,\lfloor \frac{t}{N_c} \rfloor,\cdots,\infty$, and $i \in [N_c]$.    
Each service provider $j$ is characterized by a promise quotient $s_j(t) \in [0,1]$, which is indicative of how good the quality of service provided by the $j^{th}$ provider at the event of its sale at interaction count $t$ is, where $j\in [N_p]$. Further, the service providers have a limited stock of services they sell, characterized by the maximum number of sales/purchases a service provider $j$ can undergo, denoted by $n_j^{max}$. Let $n_{t,j}$ denote the total number of sales by the $j^{th}$ provider until the interaction count $t$ since it became active. When the sales hit the threshold value $n_j^{max}$ for the $j^{th}$ service provider, the $j^{th}$ provider becomes idle or unavailable to the consumers for the next $\tau_{r}$ interaction steps; this duration serves as the total number of interaction counts it takes to refill the stock, after which the $j^{th}$ provider becomes active again. The observers are agents that observe the trade between the providers and the consumers, i.e., they observe the promise quotient of a sale/purchase. Based on these observations, the goal of the observers is to recommend high-quality service providers to consumers.
        
This paper considers a simplified model for the promise quotient capable of simulating various service provider behaviors, ranging from stable to highly volatile that we observe in the e-commerce world \cite{Liu}.  The model is described as follows:

\begin{equation} \label{eq01}
s_j(t) = \left\{
        \begin{array}{ll}
            1 & :\quad \textit{with prob.} \quad p_j(t) \\
            0 & :\quad \textit{with prob.} \quad 1-p_j(t)
        \end{array}
    \right.
\end{equation}

The DOL3 algorithm consists of three phases:
\textbf{Periodic Reset Phase:} Assists in frequent forgetting and rapid learning;
\textbf{Communication Phase:} Shares the scores among the neighbours;
\textbf{Trust Fusion Phase:} Calculates the weighted trust score based on the scores received;
\textbf{Learning Phase:} Updates the trust weights using multiplicative exponential weights update scheme.

The details of these phases are explained in Appendix \ref{phases}. 

\subsection{The DOL3 algorithm}
For the $i^{th}$ observer, $\forall i \in [N_o]$, the DOL3 algorithm involves learning the local trust weights $\hat{w}_{ij}(t)$, $\forall j \in \Omega_i$, and social trust weights $\hat{\alpha}_{lj}^i(t)$, $\forall l \in \Lambda_i$ and $\forall j \in \Omega_i \cup (\cup_{\forall l\in \Lambda_i} \Omega_l)$, which are initialized to $1$ at $t=1$, i.e., $\hat{w}_{ij}(1) = 1$ and $\hat{\alpha}_{lj}^i(1) = 1$. The local trust weight $\hat{w}_{ij}(t)$ represents the degree of trust the $i^{th}$ observer puts on the $j^{th}$ service provider which is directly connected to it as per the interaction network $G$, $\forall j \in \Omega_i$. On the other hand, the social trust weight $\hat{\alpha}_{lj}^i(t)$ represents the degree of trust the $i^{th}$ observer puts on the $l^{th}$ observer (which is directly connected to the $i^{th}$ observer, $l \in \Lambda_i$) concerning the $j^{th}$ provider's quality of service, where $ j \in \Omega_i \cup (\cup_{\forall l\in \Lambda_i} \Omega_l)$, i.e., $j^{th}$ provider is either directly interacting with the $i^{th}$ observer or the $l^{th}$ observers that are neighbors of the $i^{th}$ observer as per the interaction network $G$, $\forall l \in \Lambda_i$, or both.   

Note that the conditions in equation (\ref{eq03}) imply: if $j^{th}$ provider is observed by both the $i^{th}$ and the $l^{th}$ observer, then the social trust weight $\hat{\alpha}_{lj}^i(t+1)$ decreases if there is a mismatch between the $i^{th}$ observer's local trust score $\hat{w}_{ij}(t)$ and the $l^{th}$ observer's local trust score $\hat{w}_{lj}(t)$ since the $i^{th}$ observer would always consider its first-hand observations to be the ground-truth. Whereas, if the $j^{th}$ provider is only observed by the $l^{th}$ observer, the $i^{th}$ observer updates the associated social trust score based on the blind-trust factor $\epsilon_{\textit{trst},l}^i$ which is tuned based on how much faith / trust the $i^{th}$ observer can have on its neighboring observers in the interaction network $G$.

The trust weight $\hat{\alpha}_{lj}^i(t)$ is updated, which indicates how much trust the $i^{th}$ observer has on the $l^{th}$ neighboring observer (as per the interaction network $G$) for the trust score information on the $j^{th}$ service provider, as follows, $\forall j \in \Omega_i \cup (\cup_{\forall l\in \Lambda_i} \Omega_l)$, and $\forall l \in \Lambda_i \cup \{i\}$:
\begin{equation} \label{eq03}
\hat{\alpha}_{lj}^i(t+1) = \left\{
        \begin{array}{ll}
            \epsilon_{\textit{trst},l}^i : (l=i \wedge j\in \Omega_i) \vee (j \in (\Omega_i \cup \Omega_l)\backslash \Omega_i) & \\
            (\hat{\alpha}_{lj}^i(t))^{\gamma} \exp{(-\eta_{\alpha} |\hat{w}_{ij}(t) - \hat{w}_{lj}(t)|)} : & \\
          \quad \quad \quad \quad \quad \quad \quad \quad \quad (l\neq i) \wedge  (j \in \Omega_i \cap \Omega_l) & \\
            0 : \textit{otherwise} &
        \end{array}
    \right.
\end{equation}
and
\begin{equation} \label{eq03.5}
 \alpha_{lj}^i(t+1) = \frac{\hat{\alpha}_{lj}^i(t+1)}{\sum_{l' \in \Lambda_i \cup \{i\}} \hat{\alpha}_{l'j}^i(t+1)}
\end{equation}
where $\gamma \in (0,1]$ is the discount factor, and $\eta_{\alpha} > 0$ is the learning-rate parameter. In equation (\ref{eq03}), the first condition represents the case in which either $l = i$ and the $j^{th}$ provider is being observed by $i^{th}$ observer itself, or the $j^{th}$ provider is being observed by the $l^{th}$ observer but not the $i^{th}$ observer. The second condition is valid when $l \neq i$ and the $j^{th}$ provider is being observed by both the $i^{th}$ and $j^{th}$ observers. 

\section{Performance evaluation}
The idea of Agent Reputation and Trust (ART) testbed \cite{Kerr} which is being used for agent trust- and reputation-related technologies is extended further to simulate the real-life citizen-centric scenario of multi-agent systems interaction, which usually includes a lot of complex interactions that result in open and non-stationary environments, which is the main motivation behind developing a simulator to generate uncertainty in the data and include dynamic agents with random behaviors. The evaluation also involves simulating the conditions of how the agents are connected through the social network types - Small world, Random, and Free scale. This ensures that the models are built to scale and work across various types of networks in terms of volume, connectivity, and complexity. 

\subsection{Simulation evaluation and comparison}\label{comparison}

\begin{figure}[t]
    \centering
    \includegraphics[width=0.7\textwidth]{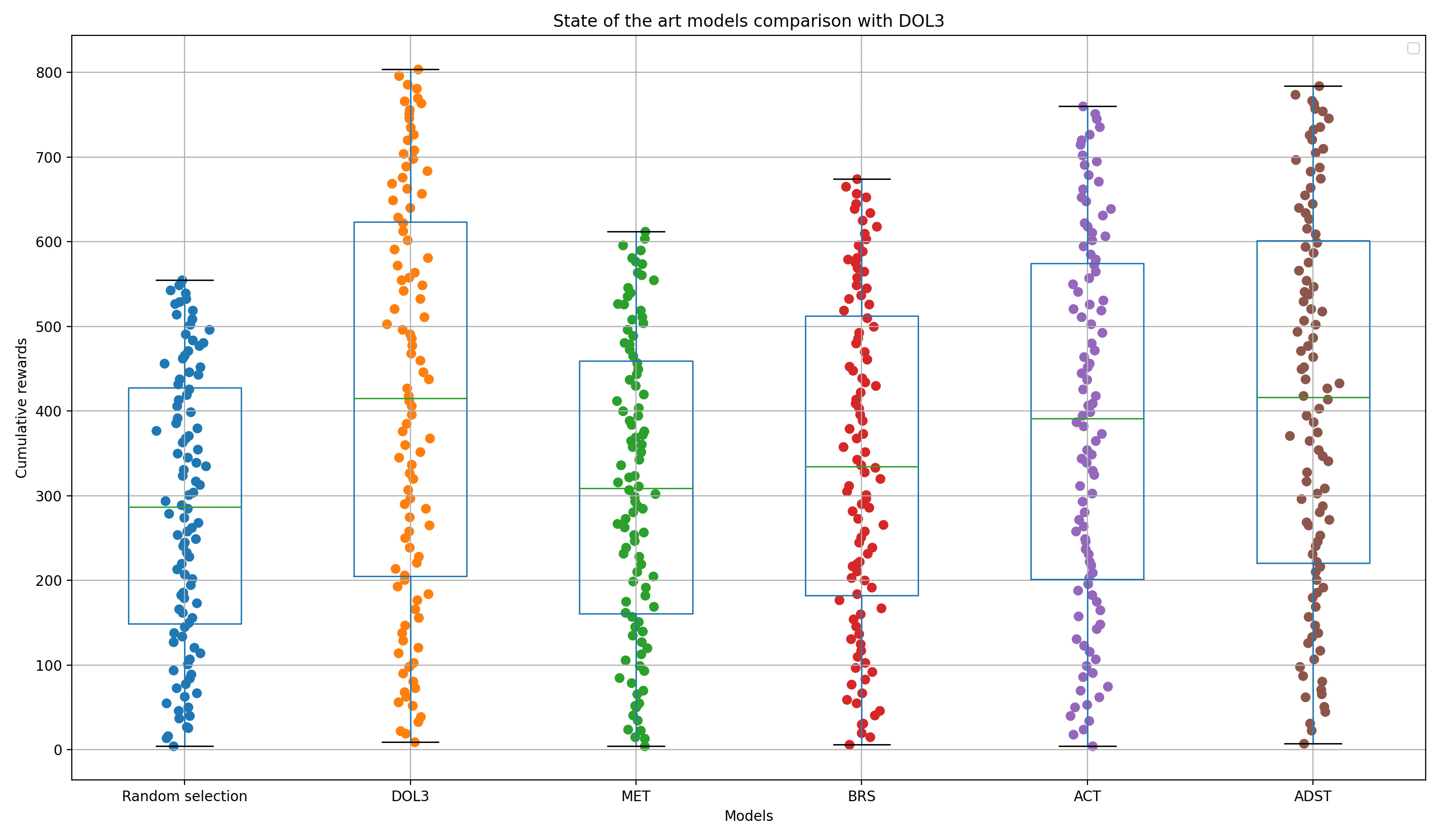}
    \caption{State-Of-the-art models comparison with DOL3 in Dynamic network}
    \label{fig:sotadol3}
\end{figure}
The top models from the baseline execution were compared with the DOL3 model in a dynamic environment with multiple Monte Carlo runs. As shown in Fig. \ref{fig:sotadol3}, the DOL3 and ADST performed better than the rest of the models. The DOL3 algorithm was configured in the simulator by changing the hyperparameters like $n_{reset}$ to an optimal value along with the discount factor ($\gamma$). The heterogeneity of the environment characterized by new providers and the deception of agents characterized by the service quality doesn't impact the speed at which the observers learn the ecosystem. From the various simulation runs, it is evident that DOL3 outperforms the other SOTA models in 90\% of the cases. 

\section{Conclusion}
With the advancement in e-commerce multi-agent architectures, Trust and Reputation Assessment plays a vital role in ensuring the quality of services. The DOL3 algorithm assists in assessing the trust of the provider and observers in a distributed fashion, where each observer learns to recommend trustworthy service providers to the consumers in real time via DOL3's adaptive online learning architecture. The simulation studies show that DOL3 performs substantially better than machine learning methods like MET, ACT, ADST, SPORAS and HISTOS, owing to its multi-layered online learning coupled with a weighted trust score fusion process and the information sharing among the observers. Further, DOL3's periodic reset phase handles the exploration part of the learning, which takes care of the high volatility in the environment; learned (biased) weights are forgotten and re-initialized after every $T_p$ discrete time-steps. The loss incurred due to such frequent explorations is reduced substantially because of the high convergence rate of DOL3's online learning, owing to the multiplicative exponential weights update scheme. With all the comparisons and statistical evaluations on the real world data, it is evident that DOL3 performs better than the state-of-the-art models in 90\% of the cases.

\subsection{Limitation and future work}
 The trade-off between exploration and exploitation in DOL3 needs further investigation. This paper considers all the provider-consumer interactions to be of the same context; DOL3 can be further extended to handle the different or changing contexts scenario. The current problem setting assumes that all the consumers are rational, i.e., they will agree to the observers' recommendations; the problem can be modified further to include irrational consumers as well. 
 
%
% ---- Bibliography ----
%
% BibTeX users should specify bibliography style 'splncs04'.
% References will then be sorted and formatted in the correct style.
%
% \bibliographystyle{splncs04}
% \bibliography{mybibliography}

\begin{thebibliography}{8}
\bibitem{Dorri}
Dorri, A., Kanhere, S. S., \& Jurdak, R. (2018). {\it Multi-agent systems: A survey}. Ieee Access, 6, 28573-28593.

\bibitem{Shamshirband}
Shamshirband, S., Anuar, N. B., Kiah, M. L. M., \& Patel, A. (2013). {\it An appraisal and design of a multi-agent system based cooperative wireless intrusion detection computational intelligence technique}. Engineering Applications of Artificial Intelligence, 26(9), 2105-2127.

\bibitem{M. Tomášek and J. Trelová}
M. Tomášek and J. Trelová, "An e-commerce applications based on the multi-agent system," 2012 {\it IEEE 10th International Conference on Emerging eLearning Technologies and Applications (ICETA), Star Lesn}, Slovakia, 2012, pp. 391-394.

\bibitem{Jășcanu}
Jășcanu, N., Jășcanu, V., \& Nicolau, F. (2007). {\it A new approach to E-commerce multi-agent systems}. The annals of “Dunarea de Jos “University of Galati. Fascicle III, electrotechnics, electronics, automatic control, informatics, 30, 11-18.

\bibitem{Zhang}
Zhang, K., Cao, Q., Sun, F., Wu, Y., Tao, S., Shen, H., \& Cheng, X. (2023). {\it Robust Recommender System: A Survey and Future Directions.} arXiv preprint arXiv:2309.02057.

\bibitem{Wu}
Wu, Q., Sang, Y., Wang, D., \& Lu, Z. (2023). {\it Malicious Selling Strategies in Livestream E-commerce: A Case Study of Alibaba’s Taobao and ByteDance’s TikTok}. ACM Transactions on Computer-Human Interaction, 30(3), 1-29.

\bibitem{Ramchurn et al.}
S. D. Ramchurn, D. Huynh, and N. R. Jennings, “{\it Trust in multi-agent
systems,” The Knowledge Engineering Review}, vol. 19, pp. 1–25, 2004.

\bibitem{Jan Tscheke et al.}
Tscheke, J., Mr, Attrey, A., Ms, Lesher, M., Ms, Carblanc, A., Ms, \& Ferguson, S., Ms (2018). {\it A Dynamic E-Commerce Landscape: Developments, Trends, and Business Models}. DIRECTORATE FOR SCIENCE, TECHNOLOGY AND INNOVATION COMMITTEE ON DIGITAL ECONOMY POLICY.  \url{https://one.oecd.org/document/DSTI/CDEP(2018)6/en/pdf}, pp. 62-64

\bibitem{Ehikioya}
Ehikioya, S. A., \& Zhang, C. (2018). {\it Real-time Multi-Agents Architecture for E-commerce Servers}. Int. J. Networked Distributed Comput., 6(2), 88-98.

\bibitem{Chelarescu}
Chelarescu, P. (2021). {\it Deception in social learning: A multi-agent reinforcement learning perspective}. arXiv preprint arXiv:2106.05402.

\bibitem{Cesa}
Cesa-Bianchi, Nicolo, and Gábor Lugosi. {\it Prediction, learning, and games}. Cambridge university press, 2006.

\bibitem{Dimitri}
Dimitri, G. M. (2023). {\it Is Facebook regionally a small world network?}. arXiv preprint arXiv:2301.04916.

\bibitem{Goyani}
Goyani, M., \& Chaurasiya, N. (2020). {\it A review of movie recommendation system: Limitations, Survey and Challenges}. ELCVIA: electronic letters on computer vision and image analysis, 19(3), 0018-37.

\bibitem{Zacharia}
G. Zacharia \& P. Maes, “{\it Trust management through reputation
mechanisms},” Applied Artificial Intelligence, vol. 14, pp. 881-907,
2000.

\bibitem{Liu}
Liu, X. (2007, July). {\it A multi-agent-based service-oriented architecture for inter-enterprise cooperation system}. In 2007 Second International Conference on Digital Telecommunications (ICDT'07) (pp. 22-22). IEEE.

\bibitem{YuH}
Yu, H. (2014). {\it Situation-aware trust management in multi-agent systems (Doctoral dissertation)}.

\bibitem{travos}
Teacy, W. L., Patel, J., Jennings, N. R., \& Luck, M. (2006). {\it Travos: Trust and reputation in the context of inaccurate information sources}. Autonomous Agents and Multi-Agent Systems, 12, 183-198.


\bibitem{Jiang}
Jiang, S., Zhang, J., \& Ong, Y. S. (2013, May). {\it An evolutionary model for constructing robust trust networks.} In AAMAS (Vol. 13, pp. 813-820).

\bibitem{Wang}
Wang, N., \& Wei, D. (2022). {\it An Adaptive Dempster-Shafer Theory of evidence Based Trust Model in Multiagent Systems}. Applied Sciences, 12(15), 7633.

\bibitem{kaggle}
\url{https://www.kaggle.com/datasets/rounakbanik/the-movies-dataset}

\bibitem{huynh}
Huynh, T. D. (2006). {\it Trust and reputation in open multi-agent systems} (Doctoral dissertation, University of Southampton).

\bibitem{Kerr} 
Kerr, R., \& Cohen, R., 2009. {\it An Experimental Testbed for Evaluation of Trust and Reputation Systems}. In: Ferrari, E., Li, N., Bertino, E., Karabulut, Y. (eds) Trust Management III. IFIPTM 2009. IFIP Advances in Information and Communication Technology, vol 300. Springer, Berlin, Heidelberg. \doi{10.1007/978-3-642-02056-8_16}

\bibitem{Masad}
Masad, D., \& Kazil, J. (2015, July). {\it MESA: an agent-based modeling framework}. In 14th PYTHON in Science Conference (Vol. 2015, pp. 53-60).

\bibitem{Kerr2}
Kerr, R., \& Cohen, R. (2010). {\it Treet: the trust and reputation experimentation and evaluation testbed}. Electronic Commerce Research, 10, 271-290.


\bibitem{sabater}
Sabater, J., \& Sierra, C. (2001, May). {\it REGRET: reputation in gregarious societies}. In Proceedings of the fifth international conference on Autonomous agents (pp. 194-195).

\bibitem{Balakrishnan}
Balakrishnan, V., \& Majd, E. (2013). {\it A comparative analysis of trust models for multi-agent systems}. Lecture Notes on Software Engineering, 1(2), 183.

\bibitem{Krampl}
Krampl, J. (2023). of {\it Thesis: Multi agent model of epidemics with socially. Learning}, 42(8), 1064-1077 (pp. 49-54).
\end{thebibliography}
%

\section{Appendix}

\subsection{Related work} \label{relatedwork}
While most researchers rely on contracts for provider-consumer interaction, tracking the transactions and the utility of the corresponding outcome is complex in a large, open, and dynamic environment. In a community of heterogeneous agents where policies define the characteristics of operations, the trust is bounded by the available information. Trust is established based on the past behavior of the agent, and historical events are used computationally to infer or predict future behavior. As stated by \cite{Zacharia}, there are various basic requirements based on which a trust model can be built, stated as follows:

\begin{itemize}
\item Effective trust measure by the trust model
\item Capability to handle open MAS
\item Robustness against deceptive agents
\end{itemize}

SPORAS \cite{Zacharia}, was used in eBay and Amazon by modeling users' trust centrally through rating aggregation. SPORAS does not consider some of the requirements like the domain or context of the environment and past experience in interacting with the provider. ReGret \cite{sabater} enables each agent to evaluate the reputation by themselves. However, ReGret doesn't take into account the problem of deceptive agents. DOL3 handles the above-stated requirements quite effectively through its multi-layered adaptive online learning of trust scores of the providers and the observers in a decentralized multi-agent architecture with information sharing among the observers. The other models like the Trust Computational Model (TCM) and MARSH as specified in \cite{Balakrishnan}, use situational and ontological references to compute trust. However, none of them considered the fact of newcomers or changes in the total number of agents in the environment. 

As stated in \cite{huynh}, a set of basic requirements (like Interaction trust, Role-based trust, Witness reputation, etc.) must be considered in a Trust and Reputation System. The above set of Trust and Reputation models come up with limitations like SPORAS not considering the social knowledge, HISTOS not having authority on the recommendations, and Beta Reputation System (BRS) having a cold-start problem for new agents entering the environment. Further, some of the evolutionary models that were explored for comparison with DOL3 include: TRAVOS \cite{travos}, Eigen Trust \cite{huynh}, Actor-Critic-Trust (ACT) \cite{YuH} - With bootstrap errors, Multiagent Evolutionary Trust model (MET) \cite{Jiang} - Having issues with observers' fairness not considered, Adaptive Dempster-Shafer Theory (ADST) \cite{Wang} assumes that there is no partial treatment of agents by one another. 

\subsection{Typical connectivity among the agents} \label{connectivity}

\begin{figure}[t]
    \centering
    \includegraphics[width=0.5\textwidth]{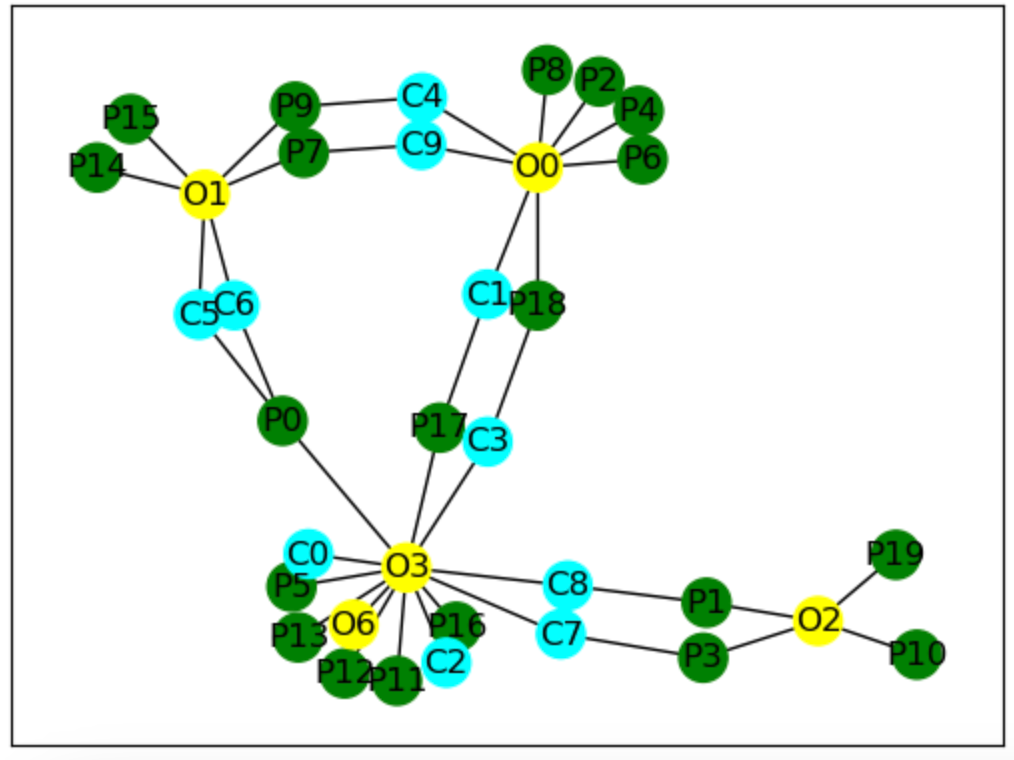}    
    \caption{The communication link among the Observer, Consumer, and Service Provider}
    \label{fig:BigPicture}
\end{figure}

In Fig. \ref{fig:BigPicture}, the observers (prefixed 'O'), consumers (prefixed 'C'), and providers (prefixed 'P') are labeled and shown as connected over an interaction network to represent their interactions (transactions, observations, and communication). The communication is also restricted to the set of agents as indicated in Fig. \ref{fig:BigPicture}.

\subsection{Phases of DOL3} \label{phases}

An iteration of the DOL3 algorithm involves the following phases: 

\textbf{Periodic Reset Phase:} The trust weights $\hat{w}_{ij}(t)$, $\forall j \in \Omega_i$, and $\hat{\alpha}_{lj}^i(t)$, $\forall l \in \Lambda_i$ and $\forall j \in \Omega_i \cup (\cup_{\forall l\in \Lambda_i} \Omega_l)$, are re-initialized to $1$ after every $T_p$ interactions. This ensures that the weights do not get biased as the number of interactions increases and can handle the non-stationary nature of the service providers' behavior. This comes as a consequence of the frequent forgetting along with the rapid learning made possible due to the exponential weights update process in the learning phase.   

\textbf{Communication Phase:} as per the interaction network $G$, $i^{th}$ observer transmits the information $\{t, i, j, \hat{w}_{ij} \}_{\forall j \in \Omega_i}$, and in turn receives, from its neighboring $l^{th}$ observer, the tuples $\{t, l, j, \hat{w}_{lj}\}_{\forall j \in \Omega_l}$ as per the interaction network $G$, $\forall l \in \Lambda_i$. 

\textbf{Trust Fusion Phase:} the $i^{th}$ observer carries out a weighted fusion of trust weights $\hat{w}_{lj}$ from all its neighboring observers $l \in \Lambda_i$ along with its own trust weight $\hat{w}_{ij}$ for a particular service provider $j$, $\forall j \in \Omega_i \cup (\cup_{\forall l\in \Lambda_i} \Omega_l)$, to obtain the $i^{th}$ observer's final trust score of the $j^{th}$ provider, $z_{ij}(t)$, as follows:

\begin{equation} \label{eq01.5}
\hat{z}_{ij}(t) = \sum_{l \in \Lambda_i \cup \{i\}} \alpha_{lj}^i(t) \hat{w}_{lj}(t) 
\end{equation}

\begin{equation} \label{eq01.5}
z_{ij}(t) = \frac{\hat{z}_{ij}(t)}{\sum_{j' \in \Omega_i \cup (\cup_{\forall l\in \Lambda_i} \Omega_l)} \hat{z}_{ij'}(t)} 
\end{equation}

\textbf{Learning Phase:} In this phase, the trust weights are updated using a multiplicative exponential weights update scheme, which is inspired by the exponentially weighted online learning forecaster \cite{Cesa}.   

The learning phase involves two learning layers; the first one is the local learning layer, in which the $i^{th}$ observer updates the local trust weights for the service providers which are its direct neighbors as per the interaction network $G$, $\forall j\in \Omega_i$, by utilizing its observations of the purchases, as follows:
\begin{equation} \label{eq02}
    \hat{w}_{ij}(t+1) = (\hat{w}_{ij}(t))^{\gamma} \exp{(\eta_{w} \sum_{k=1}^{k_{t,j}} s_j(t))}
\end{equation}

where $\gamma \in (0,1]$ is the discount factor, and $\eta_w > 0$ is the learning-rate parameter. Note that $\hat{w}_{ij}(t)$ is indicative of how good the $j^{th}$ providers' quality of service has been as observed by the $i^{th}$ observer. 

In the second learning layer, called the social learning layer, the trust weight $\hat{\alpha}_{lj}^i(t)$ is updated.

Further,    
%\begin{equation} \label{eq04}
%    \epsilon_{\textit{trst},l}^i = \left\{
%        \begin{array}{ll}
%            1 & :\quad l = i \\
%            \epsilon_{il} & :\quad l \in \Lambda_i
%        \end{array}
%    \right.
%\end{equation}
$\epsilon_{\textit{trst},l}^i$ denotes the $i^{th}$ observer's neighbor blind-trust factor for the $l^{th}$ observer, which is equal to $1$ for $l=i$ and $\epsilon_{\textit{trst},l}^i \in [0,1]$ for $l \in \Lambda_i$. The blind-trust factor $\epsilon_{\textit{trst},l}^i$ represents the degree of blind faith or trust the $i^{th}$ observer put on its neighboring $l^{th}$ observer in the interaction network $G$. The blind-trust factor $\epsilon_{\textit{trst},l}^i$ can be tuned appropriately based on either how much blind trust should be put on a neighboring observer, or to reflect such biases of an observer in real-world scenarios. 

\begin{figure}[t]
    \centering
    \includegraphics[width=0.4\textwidth]{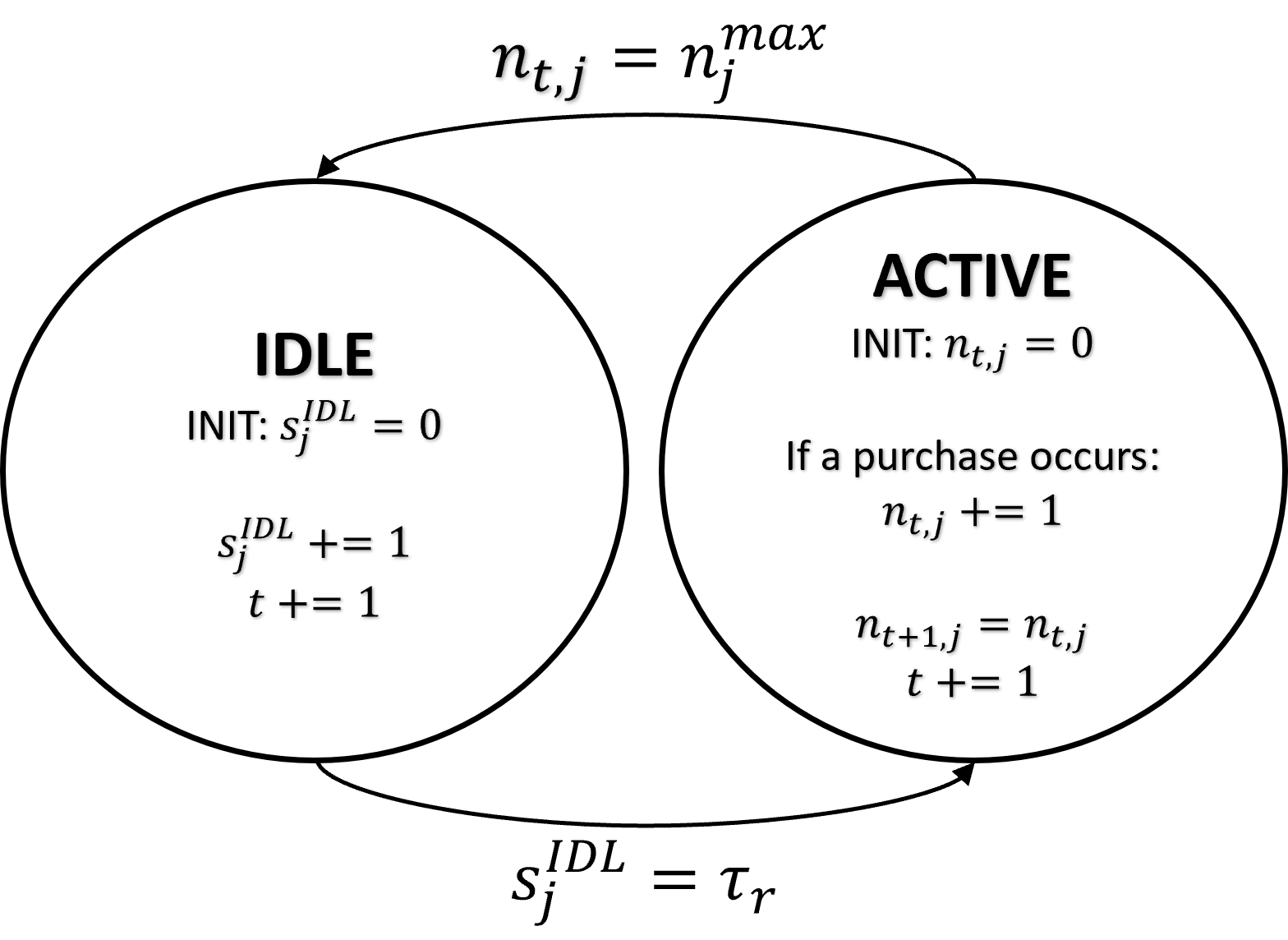}
    \caption{Active-Idle state switching model for the $j^{th}$ service provider, $\forall j\in [N_p]$; $s_j^{\textit{IDL}}$ is the step-counter in the idle state, and $n_{t,j}$ is the sales-counter of the $j^{th}$ provider in the active state.}
    \label{fig:ServiceProvider}    
\end{figure}

\subsection{Simulator setup} \label{simulator}
\begin{figure}[t]
    \centering
    \includegraphics[width=0.7\textwidth]{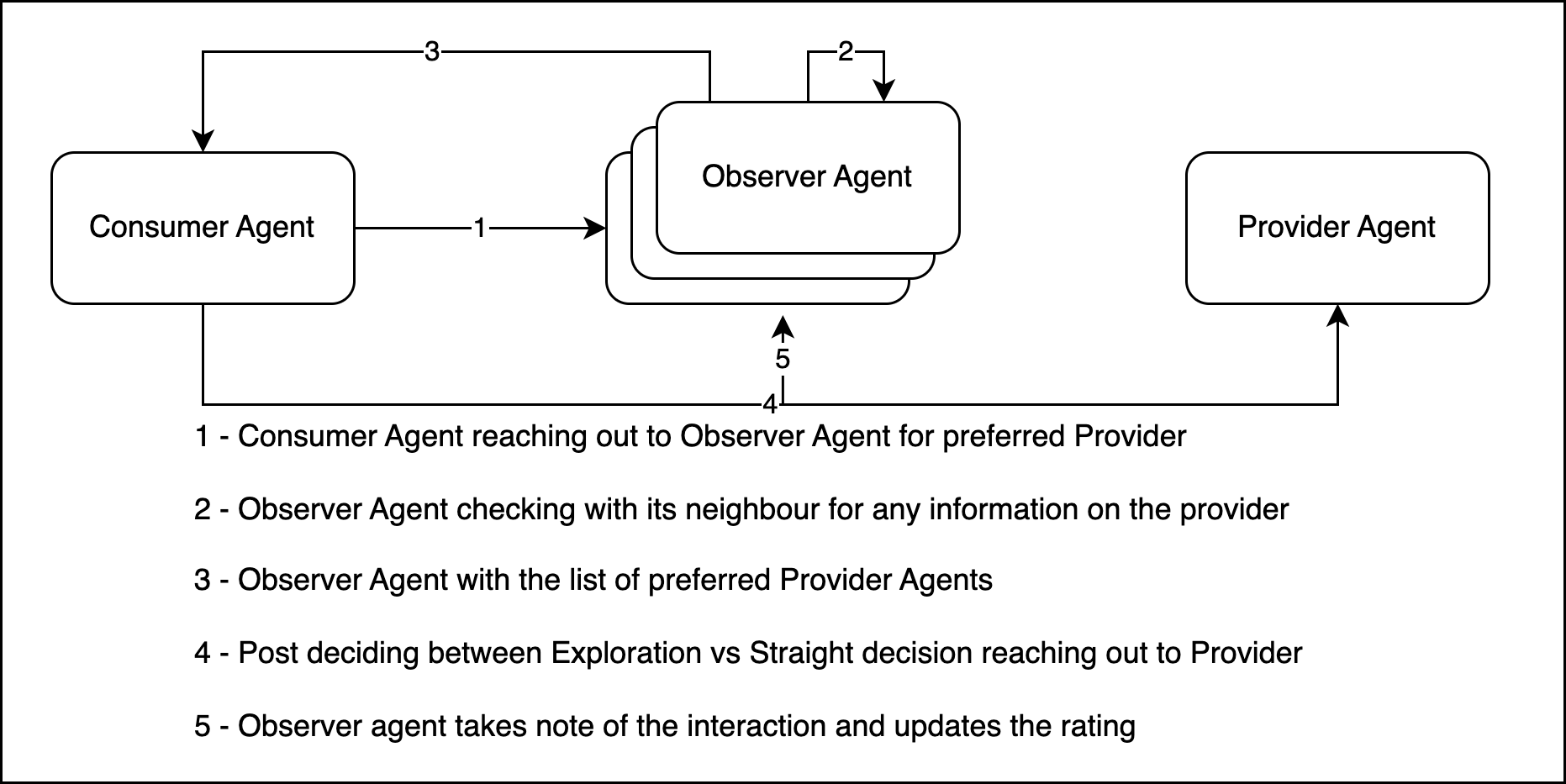}
    \caption{Simulator sequence}
    \label{fig:sequence}
\end{figure}

The simulator architecture is built on the foundation of MESA \cite{Masad}. The simulator utilizes MESA's basic components, like Agents and Schedulers, to simulate the Agents mentioned in Fig. \ref{fig:sequence} and their corresponding interactions.

\begin{figure}[t]
    \centering
    \includegraphics[width=0.7\textwidth]{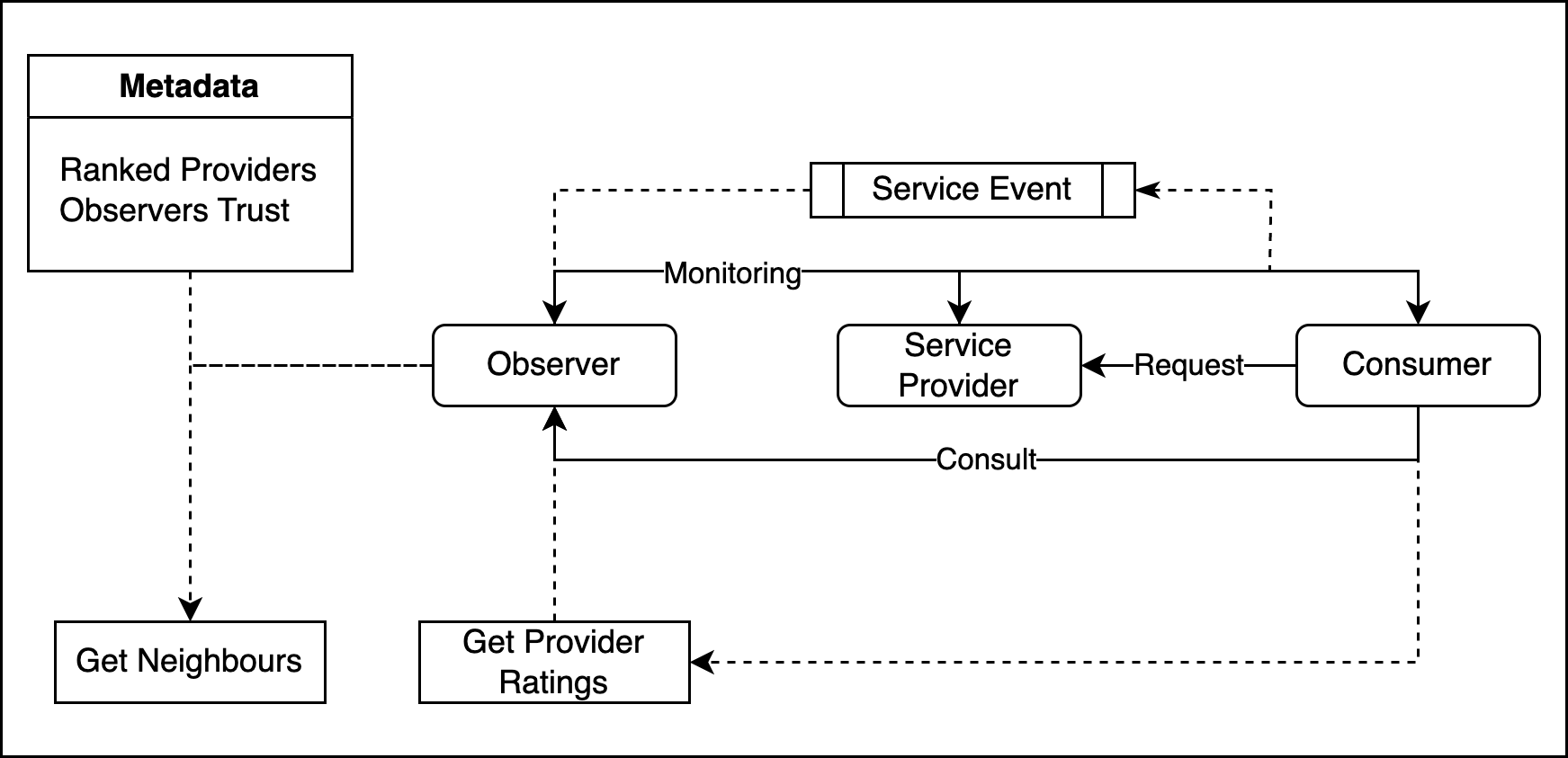}
    \caption{Simulator architecture}
    \label{fig:architecture}
\end{figure}

As illustrated in Fig. \ref{fig:architecture}, the main components of the simulator are the components like {\it Get Neighbours} and {\it Get Provider Ratings} that help in understanding the network restrictions and weighted fusion rating from all of the observers. There are several configuration capabilities that this architecture provides, allowing the evaluation of the performance of algorithms effectively. Each positive interaction is rewarded with $1$, and deceptive interaction is rewarded with $0$. The reward is randomized to introduce the non-stationary characteristic of the environment in terms of uncertainty in providers' behavior. The interactions are designed to be sequential per consumer, in the sense that only one consumer interacts with the environment at a time.  

One of the important features of the simulator is the interaction restriction among the multiple agents. This paper also shows the behavior of agents when the interactions among the agents are limited to a certain group of agents. As mentioned in \ref{problemformulation}, the consumers can receive services only from a certain set of providers. As described in Section \ref{problemformulation}, each of the providers comes with inventory and restrictions on the number of times it can serve the consumers. 

\subsection{Model comparisons} \label{comparisons}

\subsubsection{Baseline execution}
The simulation evaluation was done with the baseline version described in \ref{baseline} followed by DOL3 evaluation. Hyperparameters are used that are vital for executing the baseline and then the actual algorithm evaluation. The simulation baseline was set up by configuring the parameters mentioned in Table \ref{hyperparameters}. The randomized baseline starts by assigning random providers irrespective of the scores. The baseline, as well as the algorithm implementation, allows the consumers to either explore or exploit the ranked active providers.

\begin{figure}[t]
    \begin{subfigure}{\textwidth}
        \centering
        \includegraphics[width=\textwidth]{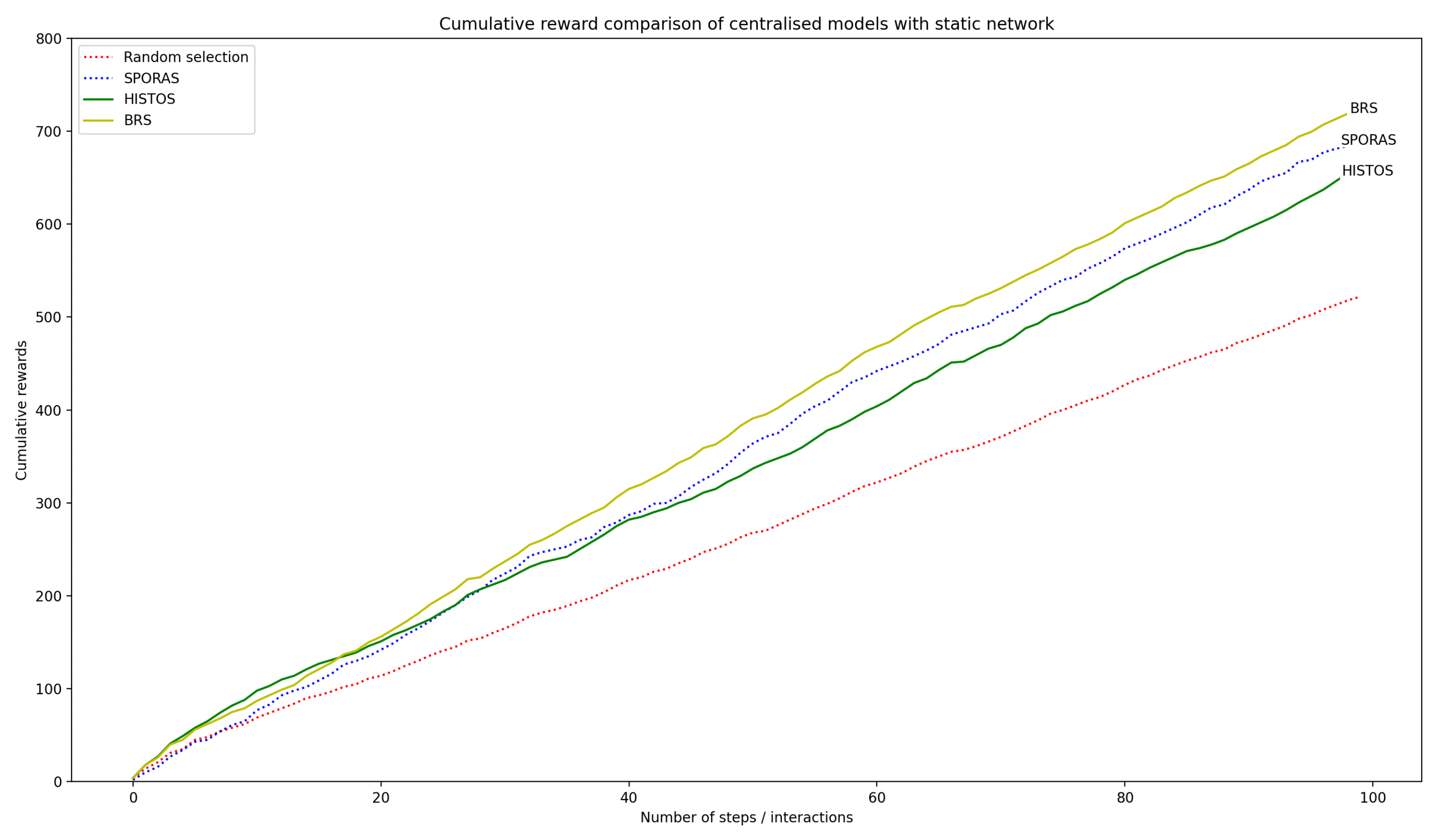}
        \caption{Static network}
        \label{fig:cbs}
    \end{subfigure}
    \begin{subfigure}{\textwidth}
        \centering
        \includegraphics[width=1\textwidth]{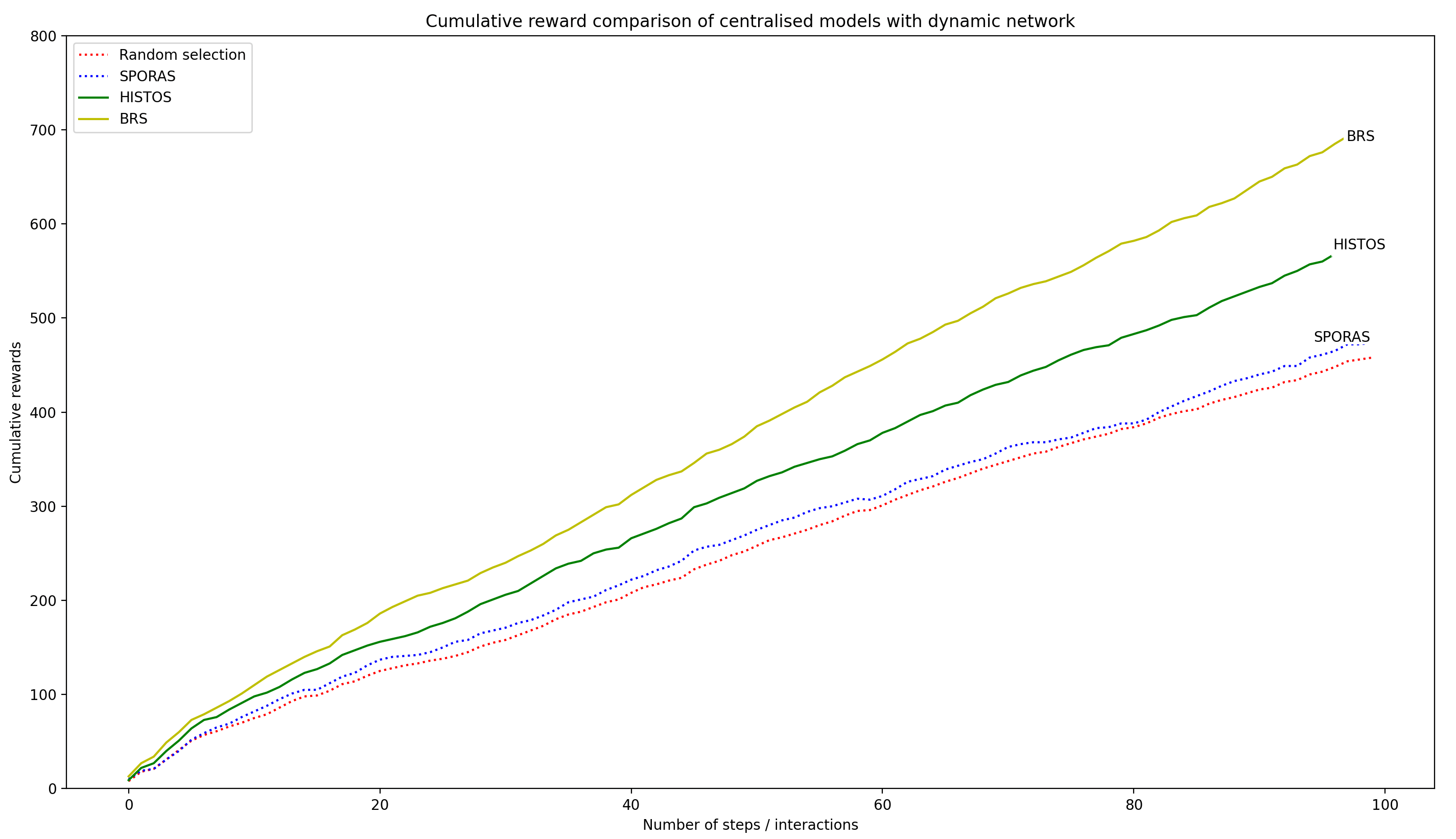}
        \caption{Dynamic network}
        \label{fig:cbd}
    \end{subfigure}
    \caption{Centralised models comparison}
\end{figure}

From the Fig. \ref{fig:cbs} and Fig. \ref{fig:cbd}, it is clear that the BRS outperforms HISTOS and SPORAS in this simulation environment on both Dynamic and Static networks. From the Decentralised models MDT, ACT, and ADST perform better across the interactions. The simulation is also considered with random observers providing expert opinions on a provider based on past interactions. The baseline clearly illustrates that the openness in the environment with the random behavior of the provider agents impacts the overall reward in the ecosystem, and the learning from the past does not add to improving the reward in the current interactions. 

The baseline helps in understanding the level of complexity the open and dynamic environment adds to the ecosystem in building Trust and Reputation. It is also clear that while past learning helps in understanding the agents' behavior, the model built out of the past interaction can not be solely relied on to determine and decide on the agents' behavior for the current interaction. With the network relation in place, the trust needs to be measured from the self-interactions as well as the opinions of the witnesses. 

\begin{figure}[t]
    \centering
    \includegraphics[width=0.7\textwidth]{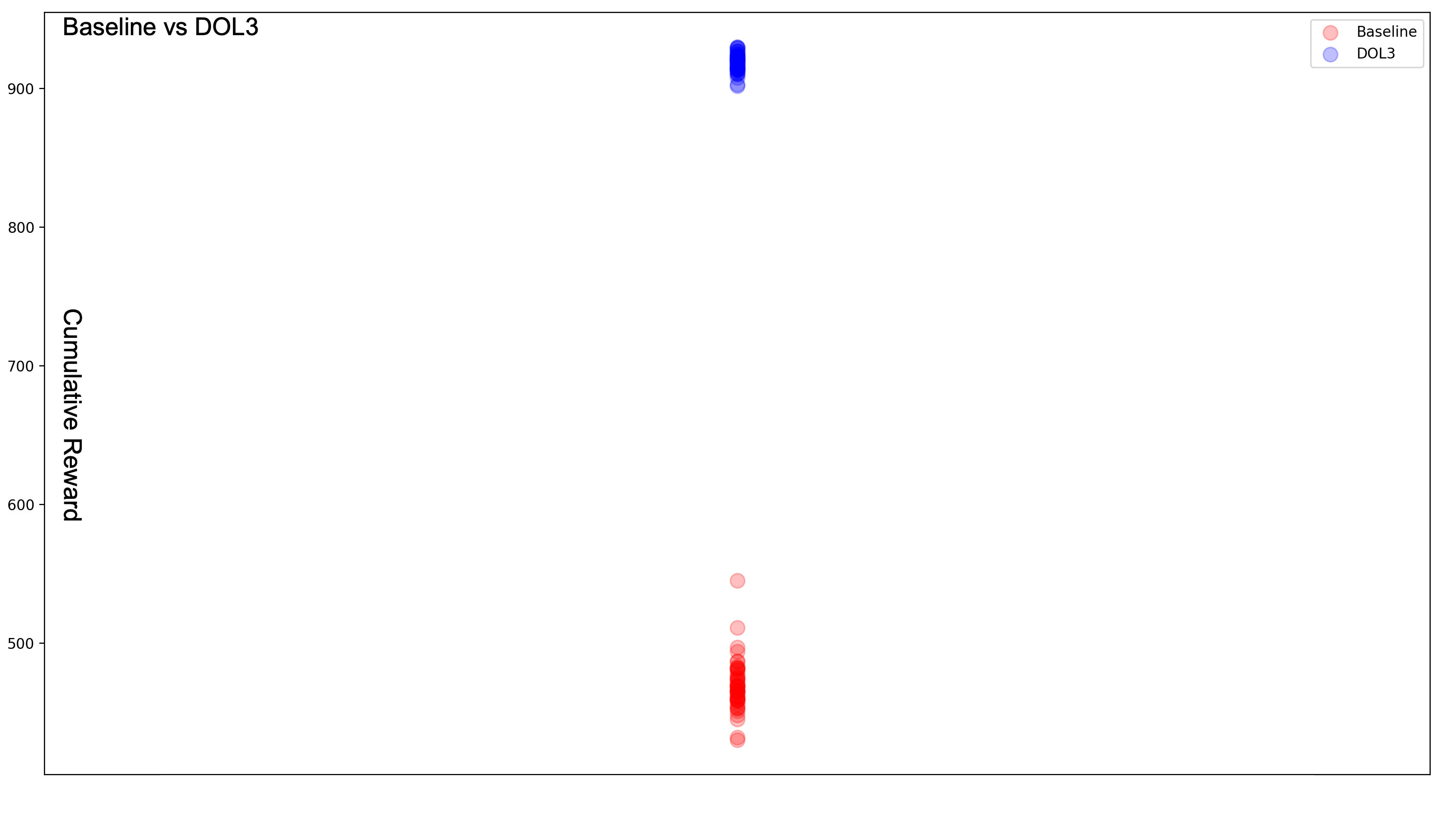}
    \caption{Simulation results of DOL3 compared with Baseline}
    \label{fig:comparison}
\end{figure}

\subsection{Baseline for evaluation} \label{baseline}
Most of the Trust and Reputation Assessment models use SPORAS as the baseline for performance \cite{Zacharia}, \cite{Kerr2}. SPORAS uses the assumption that new users start with little reputation, which builds as the services being provided increase. HISTOS was used for measuring trust in a tightly connected environment \cite{Ramchurn et al.}. We extended the baseline to contain some of the state-of-the-art models like ACT, MET, and ADST. The comparisons are done in the order mentioned below on the simulation environment built:

\begin{itemize}
\item \textbf{Centralised models in Static Network}: Comparison of the models with centralized data update protocol from the references mentioned in the above sections. 
\item \textbf{Centralised models in Dynamic Network}: The centralized models are exposed to the dynamic network where the interaction links change and the number of agents is not consistent.
\item \textbf{Decentralised models in Dynamic Network}: The decentralized models being exposed to the dynamic network.
\end{itemize}

The provider agents are ranked based on their reputation or trust scores and recommended to the consumers accordingly. The evaluation is also performed with various parameters. The following categories of baselines were considered:

\begin{itemize}
\item \textbf{Randomised Baseline}: This baseline randomly assigns reputation or prioritization scores. This lets consumers explore the agents and take a chance to be served by an agent. 

\item \textbf{Expert Opinion Baseline}: This is where the centralized observer methodology comes into the picture. The experts (observers) who know the context and have witnessed the interactions share recommendations about the providers. 

\item \textbf{Start-of-the-Art Models}: The previous State-of-the-Art models were built to measure the Trust and Reputation like that of ADST, ACT, MET, SPORAS, HISTOS, ReGret, MARSH, and TCM. 
\end{itemize}

The evaluation in this simulator consists of a combination of all the above-mentioned baselines. The baselines are customized to fit the problem statement and the characteristics of the environment considered. 

\subsubsection{Comparison of results}
We ran a Monte Carlo simulation with 100 steps split between the baseline and DOL3. The experimental result showing the cumulative reward (Sum of all the rewards per iteration) is shown in Fig. \ref{fig:comparison}. It is evident from the results that the baseline is spread on the lower bound of the rewards and is widely spread. However, the DOL3 has very little variance and spread on the upper bound. It is important to notice the variance of DOL3 showcasing that the dynamic environment doesn't impact the quality of the algorithm. 

\begin{table}
  \caption{List of hyperparameters used along with description}
  \label{hyperparameters}
  \centering
  \begin{tabular}{lll}
    \toprule
    \multicolumn{2}{c}{Hyperparameter}                   \\
    \cmidrule(r){1-2}
    Variable     & Description     & Possible Value \\
    \midrule
    $N_c$ & \# of Consumers  & $\geq$1   \\
    $N_p$     & \# of  Providers & $\geq$1      \\
    $N_o$     & \# of Observers       &  $\geq$ 1 \\
    $N$ & Total Iterations & $min(100)$ \\
    $n_{reset}$ & Every $n^{th}$ reset step& $\geq 1$ \\
    $N_{random\_stop}$ & Randomization stops & $min(10)$ \\
    $explore$ & Explore providers & ${True / False}$ \\
    $n^{max}$ & Maximum provider stock& $\geq 1$ \\
    $\eta$ & Learning rate - Observer & $\geq 1$ \\
    $\gamma$ & Discount factor & $\geq 0$ \\
    $\epsilon$ & Neighbour Blind-trust & $\geq 0$\\
    $Observer_{ndepth}$ & \# of neighbours & $ 1 - \leq (N_c - 1)$
  \end{tabular}
\end{table}

\begin{figure}
  \centering
  \begin{subfigure}{.48\textwidth}
    \includegraphics [width=1\textwidth]{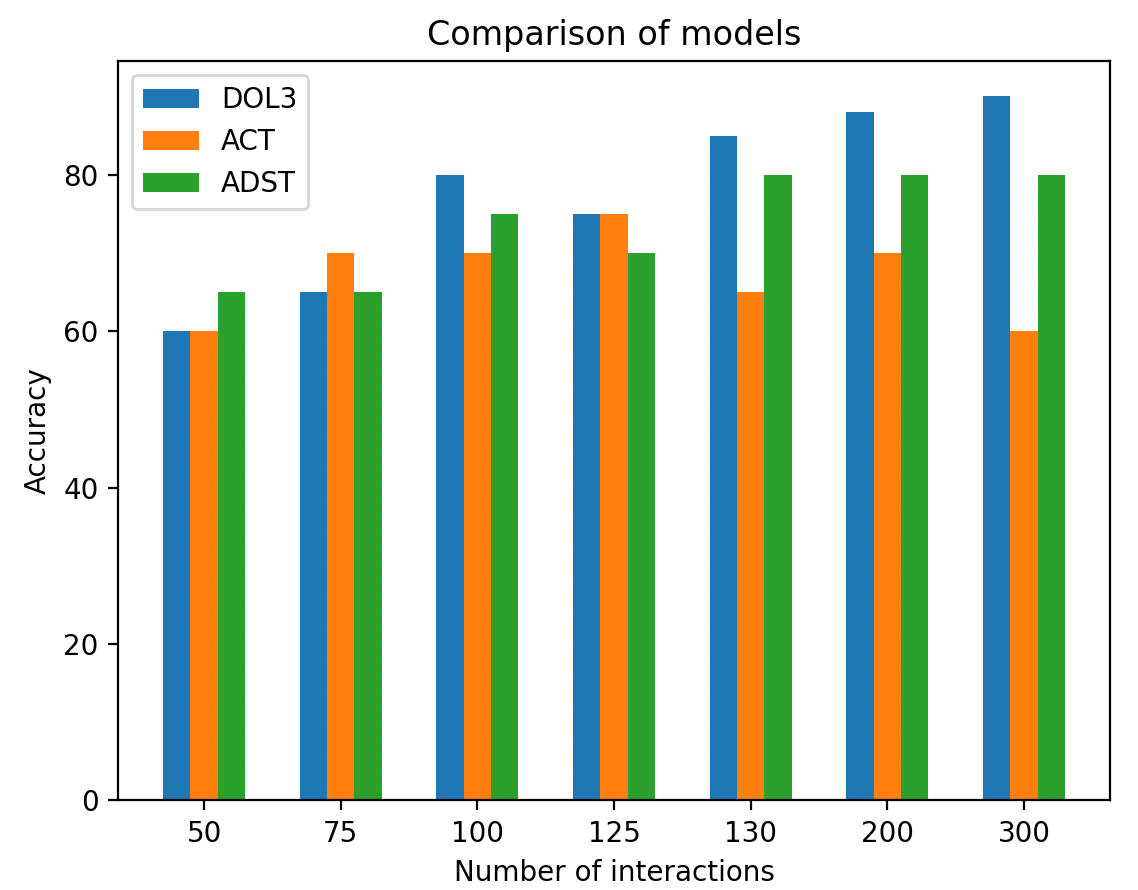}
    \caption{Interaction count}
    \label{fig:comparison_interaction}
  \end{subfigure}
  \begin{subfigure}{.48\textwidth}
    \centering
    \includegraphics [width=1\textwidth]{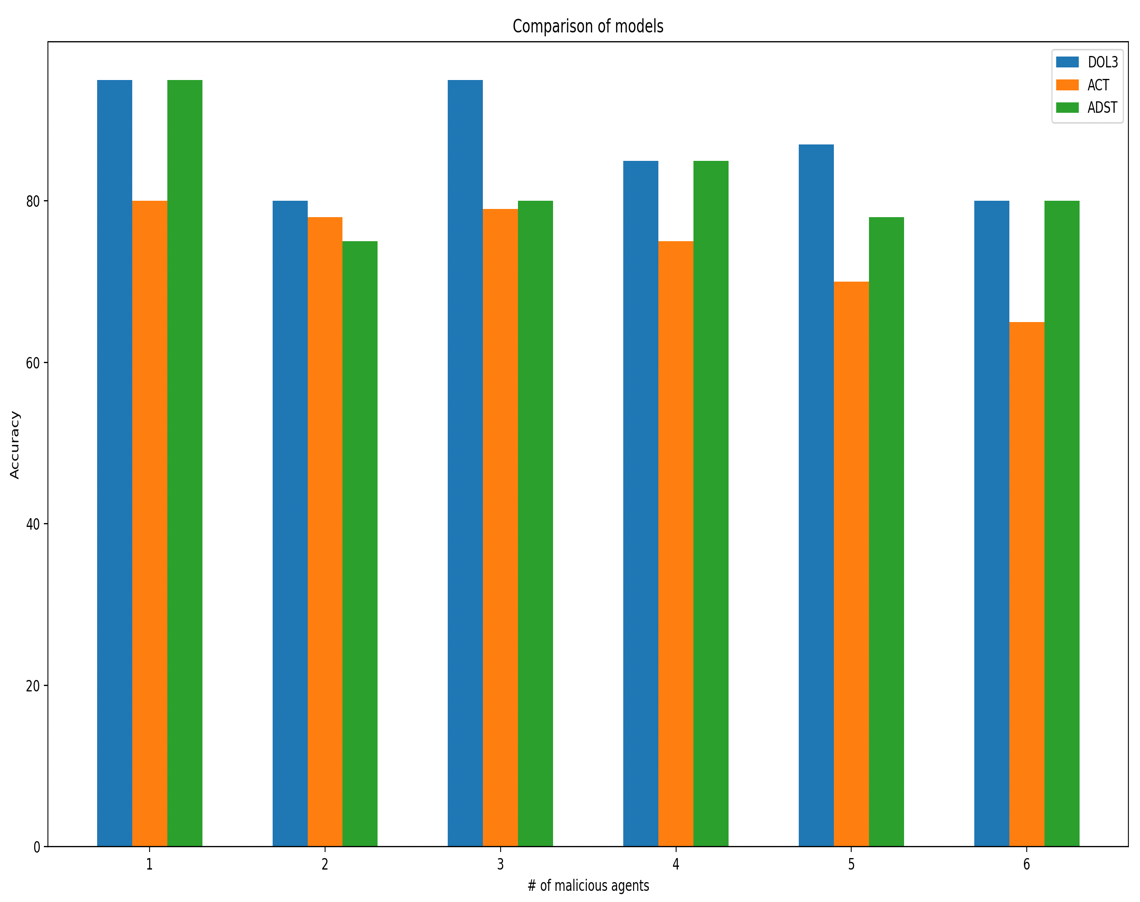}
    \caption{Number of malicious agents}
    \label{fig:comparison_malicious}
  \end{subfigure}
  \caption{Comparison of models with interaction count and malicious agents}
\end{figure}
\begin{figure}
  \centering
\end{figure}
\begin{figure}
  \centering
  \begin{subfigure}{.48\textwidth}
    \centering
    \includegraphics [width=1\textwidth]{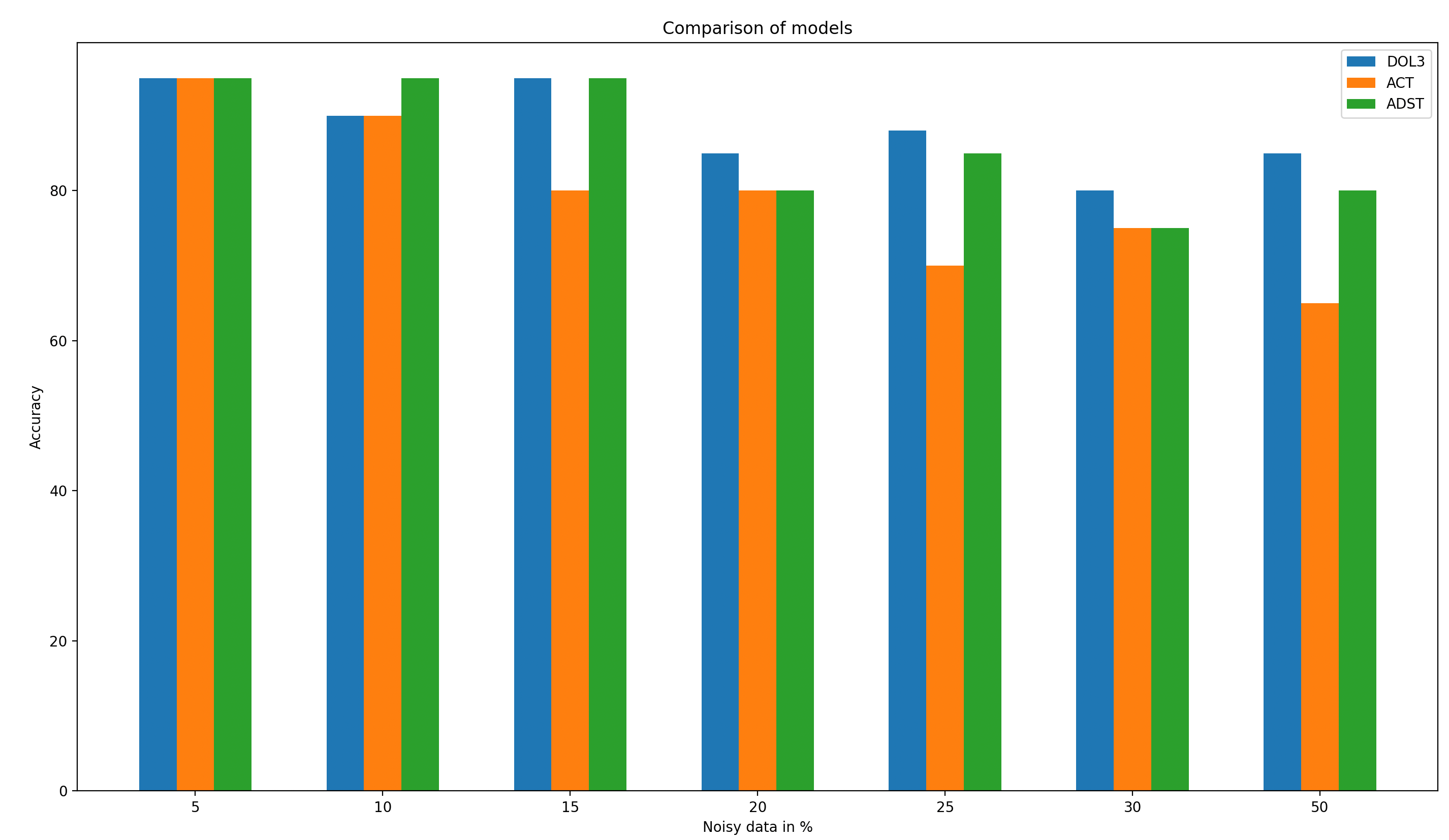}
    \caption{Sanity of data}
    \label{fig:comparison_noisy}
  \end{subfigure}
  \begin{subfigure}{.48\textwidth}
    \centering
    \includegraphics [width=1\textwidth]{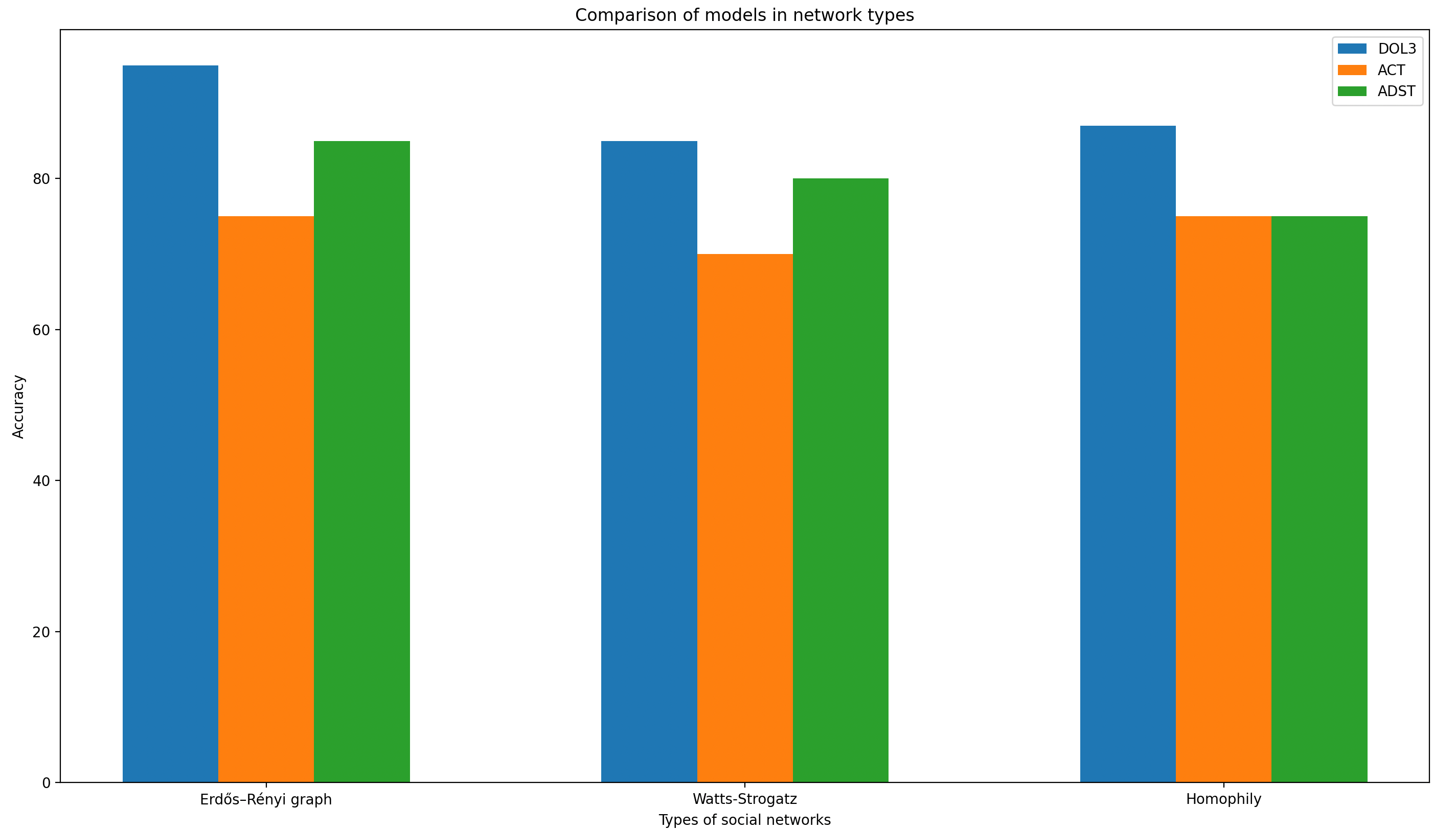}
    \caption{Type of network}
    \label{fig:comparison_types}
  \end{subfigure}
  \caption{Comparison of models based on sanity and type of network}
\end{figure}

\subsection{Statistical validation}\label{statistical}
The above simulation results help us evaluate the performance of the models against the type of network along with complexity. We applied the same against real-time data of movie recommendation system data set \cite{kaggle}. The recommendation system consisted of consumer agents (users) and service providers (recommenders) along with observers that were connected to represent various social network types like Erdős–Rényi, Watts Strogatz, and Homophily-based networks \cite{Krampl}. 

Fig. \ref{fig:comparison_interaction} shows how the models perform with the number of interaction counts. The performance or the accuracy is determined by the Root Mean Square Error (RMSE) which is given by the equation:

\begin{equation} \label{rmse}
    RMSE= \sqrt{\frac{1}{N} *(r-\hat{r})^2}  
\end{equation}

where $r$ refers to the actual rating of a movie from the data set and $\hat{r}$ refers to the rating from a recommender. 

The accuracy is given by 

\begin{equation} \label{accuracy}
    accuracy \% = \frac{1}{(1+RMSE)} * 100 
\end{equation}

The DOL3 algorithm seems to improve with the larger interaction count compared to that of other models. Similarly, Fig. \ref{fig:comparison_malicious} indicates the performance with the number of malicious agents. We artificially introduced noisy data in the data set to see how the models react when the data is corrupted. We could notice that DOL3 and ADST are more susceptible to noisy data. We could also notice from Fig. \ref{fig:comparison_types} that in the network types like that of small world and random, DOL3 performs well.  

\end{document}